\documentclass{article}
\usepackage{ijcai24}
\usepackage{multirow}
\usepackage{times}
\usepackage{soul}
\usepackage{url}
\usepackage[hidelinks]{hyperref}
\usepackage[utf8]{inputenc}
\usepackage[small]{caption}
\usepackage{graphicx}
\usepackage{amsmath}
\usepackage{amsthm}
\usepackage{booktabs}
\usepackage{algorithm}
\usepackage{algorithmic}
\usepackage[switch]{lineno}

\title{CRE-LLM: A Domain-Specific Chinese Relation Extraction Framework with Fine-tuned Large Language Model}

\author{
Zhengpeng Shi$^1$
\and
Haoran Luo$^{2}$\thanks{\ Corresponding author.}
\affiliations
$^1$College of Statistics and Mathematics, Zhejiang Gongshang University, China\\
$^2$School of Computer Science, Beijing University of Posts and Telecommunications, China\\
\emails
\texttt{shizhengpeng@outlook.com},
\texttt{luohaoran@bupt.edu.cn}
}

\begin{document}

\maketitle

\begin{abstract}
Domain-Specific Chinese Relation Extraction (DS-CRE) aims to extract relations between entities from domain-specific Chinese text. Despite the rapid development of PLMs in recent years, especially LLMs, DSCRE still faces three core challenges: complex network structure design, poor awareness, and high consumption of fine-tuning. Given the impressive performance of large language models (LLMs) in natural language processing, we propose a new framework called CRE-LLM. This framework is based on fine-tuning open-source LLMs, such as Llama-2, ChatGLM2, and Baichuan2. CRE-LLM enhances the logic-awareness and generative capabilities of the model by constructing an appropriate prompt and utilizing open-source LLMs for instruction-supervised fine-tuning. And then it directly extracts the relations of the given entities in the input textual data, which improving the CRE approach. To demonstrate the effectiveness of the proposed framework, we conducted extensive experiments on two domain-specific CRE datasets, FinRE and SanWen. The experimental results show that CRE-LLM is significantly superior and robust, achieving state-of-the-art (SOTA) performance on the FinRE dataset. This paper introduces a novel approach to domain-specific relation extraction (DSCRE) tasks that are semantically more complex by combining LLMs with triples. Our code is publicly available\footnote{\url{https://github.com/SkyuForever/CRE-LLM}}.
\end{abstract}

\section{Introduction}

\begin{figure}[t]
\centering
\includegraphics[width=8.5cm]{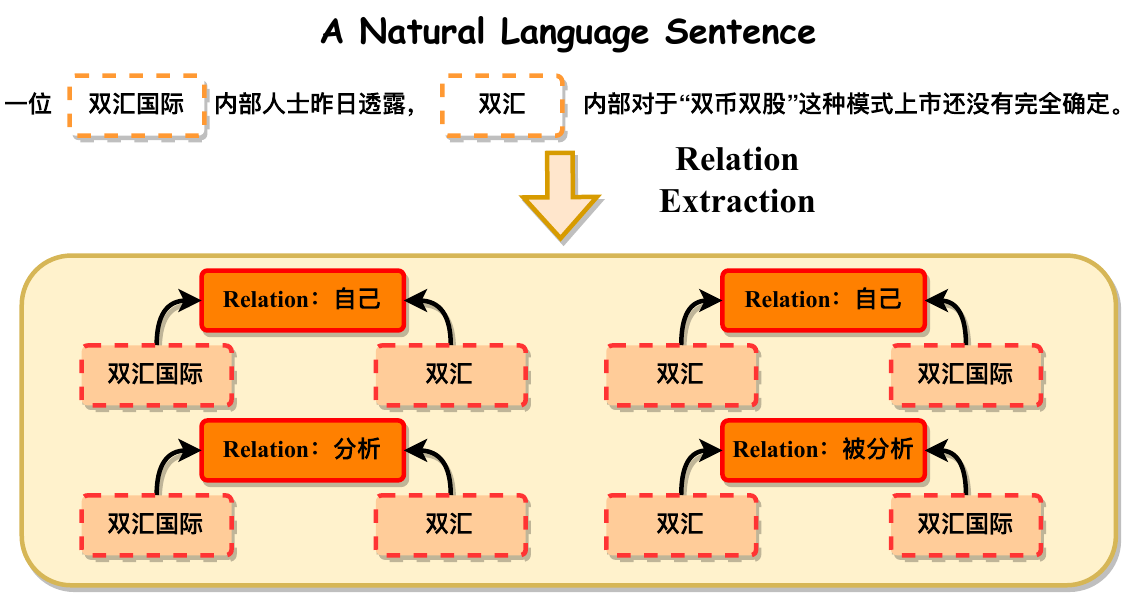}
\caption{An example of Domain-specific CRE Task.}
\label{f0}
\end{figure}
\begin{figure*}[h!t]
\centering
\includegraphics[width=16.6cm]{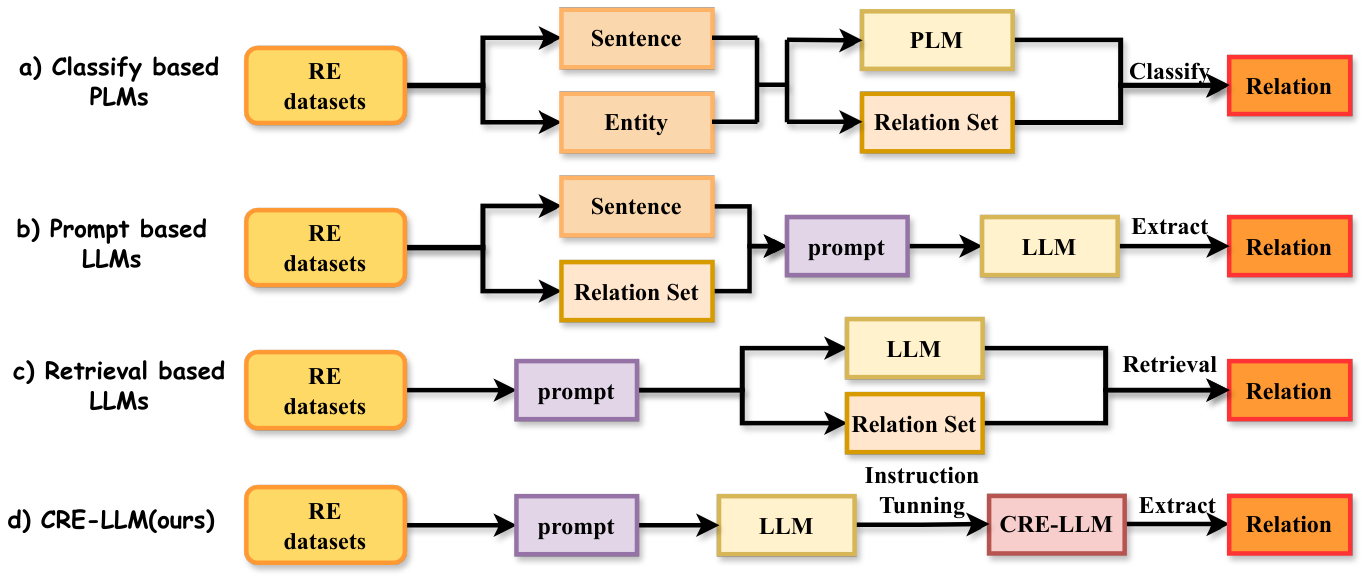}
\caption{Illustration of 4 different paradigms for solving CRE task. As shown in Figure 2a, entities and texts from the RE datasets are inputted separately into the PLM. And the PLM is combined with the Relation Set and output the relation with the highest probability as result. As shown in Figure 2b, prompts are constructed based on the texts and Relation Set from the RE dataset and input them into the LLM to generate relation. As shown in Figure 2c, the RE dataset is employed to construct the prompts and input them into the LLM to generate preliminary results, which are subsequently retrieved with the Relation Set to obtain relation extraction results. As shown in Figure 2d, our method directly utilizes a fine-tuning dataset constructed from the RE dataset to fine-tune the LLM and then generate accurate relation extraction results.}
\label{f1}
\end{figure*}
Domain-Specific Chinese Relation Extraction (DSCRE) is a crucial task in the field of Natural Language Processing (NLP). Its objective is to extract relations between given entities from domain-specific unstructured Chinese text. Examples of such relations include financial and biomedical relations. The main difficulties of this task are due to the fact that Chinese datasets specific to certain domains are mostly private, with limited informative data and resources. Additionally, dealing with the diversity of linguistic expressions and potential ambiguities in the text is a challenge. Furthermore, the limitations of the Chinese corpus and the low use of dummy words and lexemes in Chinese, which makes it challenging to extract relation between entities from domain-specific Chinese texts. A specific example as shown in Figure~\ref{f0}.

The field of deep learning has rapidly developed and achieved significant success in relation extraction tasks. Pre-trained models (shown in Figure~\ref{f1}a), such as BERT \cite{Bert} and T5 \cite{T5}, have been widely used in this area. For instance, when dealing with domain-specific Chinese relation extraction, BERT-PAGG \cite{Bert-PAGG} combines the positional information of entities with local features extracted by the PAGG module and entity vector representations outputted by BERT to achieve relation extraction. Similarly, MoVE \cite{MoVE} achieves this by dynamically learning multi-view features. LLMs can also be used for domain-specific relation extraction (shown in Figure~\ref{f1}b), such as ERNIE 3.0 \cite{ERNIE3.0} and ERNIE 3.0 TITAN \cite{ERNIE3.0TITAN}, the continuous learning semantic understanding frameworks were developed based on knowledge augmentation. It utilized self-supervised contrastive learning pre-training technique of word-mixing and self-adversarial fine-tuning with word-mixing data augmentation to improve RE tasks. Additionally, GPT-FinRE \cite{GPT-FinRE} can also be used for this purpose (shown in Figure~\ref{f1}c). It uses OpenAI models under the ICL framework and incorporates retrieval mechanisms to achieve financial relation extraction, has strong potential and performance in general.

Several approaches have been proposed for this task, resulting in significant performance improvements. However, three main challenges and difficulties remain. \textbf{(1) Complex network design is required.} Methods such as BERT-PAGG \cite{Bert-PAGG}, MoVE \cite{MoVE}, and other BERT-type pre-trained models have been proposed, which require the construction of complex networks and the combination of external and internal information to improve model performance. \textbf{(2) Poor perception is a challenge.} Direct use of large models for event extraction may not perceive internal Relations, especially in the domain-specific of Chinese text. A sentence contains semantic information from different perspectives, including words, structure and contextual semantic information, as well as domain-specific proper nouns. These difficulties can impede the understanding of the model, so direct use of the model for relation extraction often yields unsatisfactory results. \textbf{(3) Fine-tuning the model to achieve the domain-specific CRE task results in significant memory consumption.} GPT-3 \cite{GPT-3} demonstrates that expanding the pre-trained models can further exploit its potential. However, fine-tuning large-scale pre-trained models such as ERNIE 3.0 \cite{ERNIE3.0} and ERNIE 3.0 TITAN \cite{ERNIE3.0TITAN}  to achieve domain-specific Chinese relation extraction requires a large amount of memory consumption, which may be difficult for a typical team to achieve. 

To address the aforementioned challenges, we introduce CRE-LLM (shown in Figure~\ref{f1}d), a framework for extracting domain-specific Chinese relations using fine-tuned open-source large models such as Llama-2-7B \cite{llama2}, ChatGLM2-6B \cite{chatglm}, and Baichuan2-7B \cite{Baichuan2}. CRE-LLM proposes a direct and concise method for relation extraction by calling a fine-tuned open-source large model and constructing an appropriate prompt. (1) The method selects a pre-trained open-source large model and constructs an appropriate prompt to directly complete relation extraction. This eliminates the need to design and build complex network structures. Instead, the PEFT \cite{peft} framework can be used to achieve end-to-end question and answer quickly and efficiently \cite{ChatKBQA}. Additionally, larger models typically have more parameters and often yield better relation extraction results. (2) In order to enable the model to perceive and comprehend the internal relations between the given sentences and entities, we simultaneously utilize instruction-supervised fine-tuning methods. This enhances its logical perception and generation capabilities, resulting in appropriate extraction outcomes. (3) To leverage the enhanced performance of large-scale language models, we apply the PEFT \cite{peft} framework to fine-tune LLMs. This substantially decreases memory consumption and enhances training efficiency, enabling typical projects and teams to harness LLMs for domain-specific CRE tasks.

To evaluate the performance of our proposed framework, we applied it to datasets of CRE tasks from two different domains: FinRE \cite{CRE} and SanWen \cite{CRE}. The experimental results demonstrate the excellent performance of CRE-LLM on the domain-specific CRE tasks, while achieving new state-of-the-art (SOTA) performance on FinRE. We also conducted additional experiments to verify whether our relation extraction framework improves the relation extraction accuracy and efficiency. Finally, we also discuss how the insights from this framework allow us to envision future combinations of LLMs and domain-specific CRE.

\section{Related Work}
\label{Related}

\subsection{Domain-Specific RE with PLMs}
Categorical relation extraction based on pre-trained language models (PLMs) such as BERT \cite{Bert} and T5 \cite{T5} is currently one of the mainstream methods for extracting relations in specific domains. The two main pre-trained language models used in the biomedical domain are BioBERT \cite{BioBert} and PubMedBERT \cite{PubMedBERT}. These models are suitable for real-world tasks such as entity and relation extraction. In the financial domain, T5-base \cite{T5-base} is a commonly used pre-trained language model. It is important to note that technical term abbreviations should always be explained when first used. In order to improve the overall level of Chinese financial natural language processing (NLP) and promote the development of information extraction, several models have been proposed, including FinBERT \cite{FinBert} and Mengzi-BERT-base-fin \cite{Meg}. Furthermore, in the domain-specific CRE task, BERT-PAGG \cite{Bert-PAGG} has developed the PAGG module to enable the model to synthesize multiple pieces of information about entities. Additionally, MoVE \cite{MoVE} has integrated different view representations through hybrid viewpoints and expert mechanisms, which effectively filter out noisy information and improve the efficiency of model training and inference. 

This paper introduces CRE-LLM, a framework for DSCRE that deviates from the traditional classification-based approach for relation extraction and instead uses a generative method. Fine-tuned open-source LLMs are utilized to directly discern relations between given entities through the generation process. The framework aims to solve the complex network structure design problem of previous pre-trained models like BERT \cite{Bert} and T5 \cite{T5}.

\subsection{Domain-Specific RE with LLMs}
The emergence phenomenon has been demonstrated by the introduction of ChatGPT \cite{ChatGPT} and GPT-4 \cite{GPT4}, which are decoder-only Large Language Models (LLMs) with a large number of parameters. These models have shown strong performance on natural language problems, making many traditional NLP tasks easier \cite{llm}. The emergence of open-source large models such as Llama2-7B \cite{llama2}, ChatGLM2-6B \cite{chatglm}, and Baichuan2-7B \cite{Baichuan2} have made the use of LLMs more convenient. For example, GCRE-GPT \cite{GCRE} uses a generative approach to extract complex multiple comparison relations from input text. On the other hand, GPT-RE \cite{GPT-RE} focuses on entity and relation information and use gold label induced reasoning to bridge the performance gap of RE. GPT-FinRE \cite{GPT-FinRE}, which has strong potential to implement financial relation extraction using OpenAI models and combining retrieval mechanisms under the ICL framework. Therefore, leveraging powerful semantic parsing capabilities of the LLMs to design and implement relation extraction through multiple rounds of dialogue or retrieval is a promising approach.

In this paper, our proposed CRE-LLM is a novel approach to relation extraction. It employs a fine-tuned open-source LLMs and leverages the powerful capabilities of LLMs in text understanding, generation and generalization. This enables the extraction of relations between specified entities in a simple and direct end-to-end method from unstructured Chinese text.

\subsection{Fine-Tuning for Large Pre-Trained Models}
With the open-source LLMs like Llama-2-7B \cite{llama2}, ChatGLM2-6B \cite{chatglm} and Baichuan2-7B \cite{Baichuan2} emerging. It can be supervised fine-tuning (SFT) by using Parameter-Efficient Fine-Tuning (PEFT) technologies \cite{peft} such as LoRA \cite{LoRA}, QLoRA \cite{QLoRA}, P-Tuning v2 \cite{P-Tuing}, and Freeze \cite{Freeze}, enhancing the capabilities of LLMs for specific tasks. For instance, ERNIE 3.0 \cite{ERNIE3.0} and ERNIE 3.0 TITAN \cite{ERNIE3.0TITAN} achieved significant performance improvements on domain-specific CRE datasets through full fine-tuning. Additionally, the fine-tuning approach was used to construct the BBT-FinT5 \cite{BBT-Fin} model based on the T5 model, which promotes the development of natural language processing (NLP) in Chinese finance. In conclusion, the implementation of domain-specific CRE tasks through fine-tuning LLMs is highly effective and significant. This enables more general projects and teams to deploy implementations. 

The CRE-LLM integrates the semantic parsing capability of LLMs with the advantages of instruction-supervised fine-tuning. This provides convenience and possibility for LLMs to be deployed in DSCRE tasks.
\begin{figure*}[h!t]
\centering
\includegraphics[width=17.2cm]{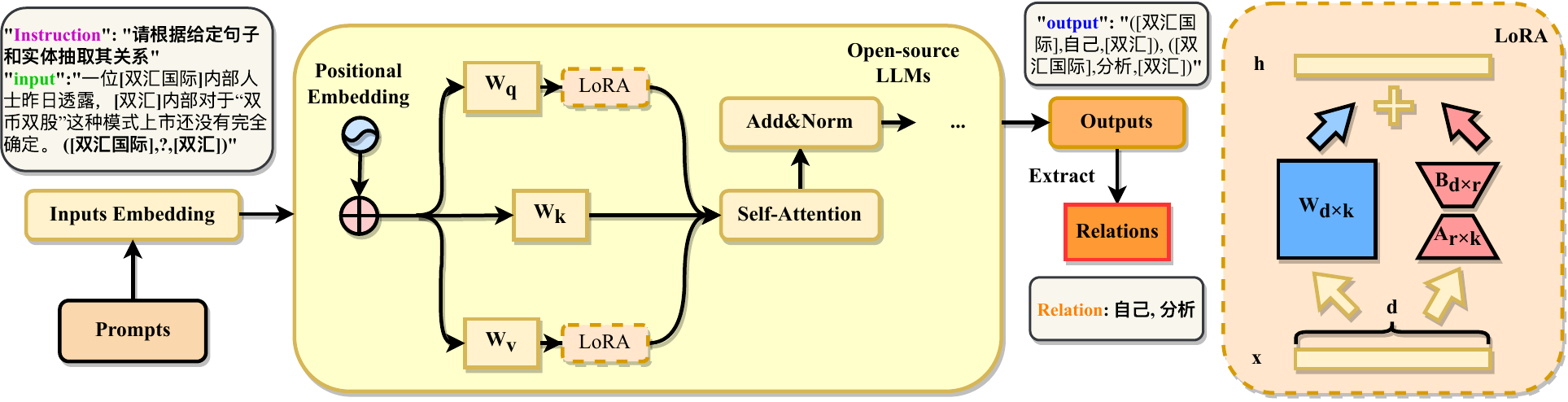}
\caption{The overview of CRE-LLM for domain-specific Chinese relation extraction method with supervised fine-tuned LLMs by using Parameter-Efficient Fine-Tuning (PEFT) technologies (e.g. LoRA).}
\label{f2}
\end{figure*}
\section{Preliminaries}

\subsection{Problem Definition}
Given an input sentence of $n$ words $s = \{x_1, x_2, ..., x_n\}$, an entity e is a consecutive span of words where $e = \{xi, xi+1, ... , x_j\}$, $i, j\in (1, ..., n)$. For each sentence s, the output of the CRE-LLM is a set of facts where each fact consists of a relation triplet. A relation triplet consists of the relation $r \in R$ between head entity $e_{head}$ and tail entity $e_{tail}$ where R is the predefined set of relation labels. Hence, a relation triplet has three components: $(e_{head}, r, e_{tail})$. A specific example is shown in Figure~\ref{f2}.
\subsection{Parameter-Efficient Fine-Tuning Methods}
The current trend in pre-trained language models (PLMs), represented by models like ChatGPT \cite{ChatGPT}, is towards increasing scale. This leads to a growing cost for full fine-tuning. PEFT \cite{peft} methods address this challenge by freezing the majority of model parameters and fine-tuning only a small or additional set of parameters. This approach achieves comparable performance to full fine-tuning while significantly reducing the overall fine-tuning costs. LoRA \cite{LoRA} , as a form of PEFT \cite{peft}, which employs low-rank approximation, introducing low-rank matrix modules labeled as A and B. This approach helps to reduce memory consumption during fine-tuning of LLMs by minimizing changes in weights associated with the parameters of the model. For instance, considering the weight matrix $W_0$ of a pre-trained model, its update is constrained through low-rank decomposition, involving training only $B$ and $A$, where $B \in R_{d\times r}, A\in R{r\times k}, r\ll min(d,k)$. The specific update is expressed as follows:
\begin{align}
    h = &  W_0x + \Delta Wx = W_0x+BAx.
\end{align}%
Therefore, LoRA \cite{LoRA} framework offers the following advantages: generalization comparable to full fine-tuning, no additional inference latency, and reduced consumption of memory and storage resources.

\section{Methodology}

In this section, we first give a brief overview of CRE-LLM and introduced by the following three main sections: Instruction design, Efficient Fine-Tuning on LLMs and Discussion of Methods. We explain how the fine-tuning dataset is constructed, why we use the PEFT \cite{peft} framework to fine-tune LLMs and comparison of CRE-LLM to other methods. 

\subsection{Overview of CRE-LLM}

CRE-LLM is a framework for domain-specific CRE based on generative question-answering, utilizing open-source LLMs fine-tuned with instruction supervision. Firstly, CRE-LLM involves constructing effective instructions based on natural language and specified entities in CRE datasets. Simultaneously, it sets up reasonable input and output configurations to enable the model to better understand and accomplish the task. Subsequently, leverage PEFT \cite{peft} framework to achieve efficient fine-tuning of LLMs. And then, from the given Chinese text data, the fine-tuned LLMs are employed to generate inference results in the form of triplets by means of prompts. Finally, the framework employs a direct extraction process to derive the relations from the generated triplets, thereby elucidating the relations between the specified entities in the Chinese text. The detailed framework description is illustrated in Figure~\ref{f2}.

\subsection{Instruction Design}

To construct an instruction fine-tuning dataset, we need to design instruction, input and output. In this context, we set the ``instruction" for inputting LLMs as ``Please extract the relation based on the given sentence and entities." And the entities in each instance of the relation extraction dataset (e.g., ``Shuanghui") are labeled with the given entities using ``[]" (e.g., ``[Shuanghui]") to help LLMs understand the meaning of the concept referred to by the term ``given entities" and better focus on and comprehend the relation between the given entities.

Subsequently, to facilitate LLMs understanding of the task requirements and ensure the accuracy of inferring relation extraction, we append a triplet and the format is that $([e_{head}], ?, [e_{tail}])$ (e.g., ``([Shuanghui International], ? ,[Shuanghui])"), where ``?" represents the relation between the given entities. The specific format of the instance data is detailed in Figure \ref{f2}. Combining the relations between given entities, we construct triplets (e.g., ``([Shuanghui International], analysis ,[Shuanghui])") as both ``input" and ``output," and finally, the previously constructed instruction is added to complete an instance in the fine-tuning dataset.

Additionally, if multiple relation exist between entities in a sentence, we list all relation in the inference results, separated by ``," (e.g.``([Shuanghui International], self, [Shuanghui]), ([Shuanghui], analysis, [Shuanghui International])"). Following this structured process, the instruction fine-tuning training dataset for open-source LLMs is constructed.

\subsection{Efficient Fine-Tuning on LLMs}
To reduce the substantial cost of fine-tuning LLMs, overcome the limitations of the available information resources for domain-specific CRE tasks and ensure that LLMs can generate standardized relation extraction results, it is necessary to implement a solution that addresses these issues. The CRE-LLM employs the PEFT \cite{peft} framework to address these challenges and minimize the number of fine-tuning parameters.

The CRE-LLM allows switching between all the aforementioned fine-tuning methods and open-source LLMs, such as Llama-2-7B \cite{llama2}, ChatGLM2-6B \cite{chatglm}, and Baichuan2-7B \cite{Baichuan2}. As shown in Figure 3, for these large-parameter-only decoder LLMs, CRE-LLM adopts PEFT \cite{peft} technology. It fine-tunes the $Q$ and $V$ parts of the input in the GQA section using LoRA \cite{LoRA}, adds them to the $K$ part, and then calculates the attention using the following formula:
\begin{align}
    Attention(Q,K,V) = &  softmax(\frac{QK^T}{\sqrt{d_k}})V.
\end{align}%
Then, through a series of network architecture layers, the relations of the given entities in the input text are generated. The underlying mechanism can be outlined by the following formula:
\begin{align}
    p_\theta(\mathcal{Y}|\mathcal{X},\mathcal{P}) = &  \prod_{i=1}^m p_\theta(y_i|\mathcal{X},\mathcal{P},y<i),
\end{align}%
where $\mathcal{X}=[x_1, x_2, ..., x_n]$ is the text sequence to be extracted, $\mathcal{Y}=[y_1, y_2, ..., y_m]$ is the target sequence, and $\mathcal{P}$ is the prompt. Through PEFT \cite{peft} framework, the problem of poor internal perceptual capabilities in general open-source LLMs is addressed. It simultaneously enhances the generation understanding and generalization abilities of LLMs with respect to texts in this domain specific. This can be widely applied to domain-specific CRE tasks, generating more accurate results.
\subsection{Discussion of Methods}

Based on the explanation and process description of the CRE-LLM method above, let's discuss and compare it with existing CRE methods:

\textbf{(1) Comparison with Classify based PLMs.} The PLMs, such as BERT \cite{Bert}, T5 \cite{T5}, ERNIE 3.0 \cite{ERNIE3.0} and ERNIE 3.0 TITAN \cite{ERNIE3.0TITAN} primarily implement CRE using classification-based methods. In contrast, the CRE-LLM utilizes PEFT framework to fine-tune LLMs and employs a generative approach for CRE.

\textbf{(2) Comparison with Classify-then-Extract based LLMs.} 
The direct invocation of LLMs, such as ChatGPT \cite{ChatGPT} and GPT4 \cite{GPT4} involves constructing a prompt based on the Classify-then-Extract approach. This prompt must include the Relation Set as options for the relation between the given entities. In contrast, CRE-LLM employs a fine-tuned LLM that has incorporated the knowledge of the Relation Set, eliminating the need for an excessively lengthy prompt while achieving the domain-specific CRE task.

\textbf{(3) Comparison with Generate-then-Retrieval based LLMs.} Regarding the Generate-then-Retrieval method, LLMs generate the relation between given entities directly and then align it with the Relation Set through retrieval. In contrast, CRE-LLM does not require additional retrieval alignment with the Relation Set. It can directly infer and generate more accurate relation extraction results.

\section{Experiments}
This section covers experimental configurations, results, and analysis. We address the following research questions (RQs): 
\textbf{RQ1}: Is CRE-LLM superior to other CRE methods? 
\textbf{RQ2}: Is fine-tuning of LLMs effective? 
\textbf{RQ3}: Is the design of the prompts for LLMs reasonable? 
\textbf{RQ4}: Why use fine-tuned open-source LLMs instead of directly invoking ChatGPT for CRE? 
\textbf{RQ5}: Does LoRA fine-tuning reduce GPU memory consumption and environmental configuration requirements, and does it improve training efficiency? 
\textbf{RQ6}: How about error analysis?

\subsection{Experimental Setup}
\textbf{Datasets.} All experiments were conducted on two Chinese datasets from different domains: FinRE \cite{CRE}, based on 2647 financial news articles from Sina Finance, containing 44 distinguished relations, including the special relation ``NA" indicating no relation between the marked entity pairs. The dataset was split into 26,971, 2,977, and 7,453 relation extraction instances for training, validation and testing respectively. SanWen \cite{CRE} is based on 837 Chinese literary works, comprising 10 distinguishable relations, also including the special relation ``NA." It was respectively divided into 17,227, 1,793 and 2,220 relation extraction instances for training, validation, and testing.

\textbf{Baselines.} CRE-LLM was compared with several CRE baselines, including ERNIE 3.0 TITAN \cite{ERNIE3.0TITAN}, Bert-PAGG \cite{Bert-PAGG}, MoVE \cite{MoVE}, and other CRE methods mentioned in Section \ref{Related}.

\textbf{Evaluation Metrics.} Following previous work \cite{ERNIE2.0,ERNIE3.0,ERNIE3.0TITAN,BBT-Fin,Bert-PAGG,MoVE}, we used Precision, Recall, and F1-Score to evaluate the performance of methods.

\textbf{Hyperparameters and Environment.} For the FinRE dataset \cite{CRE}, we fine-tuned the LLM for 5 epochs with a learning rate of 5e-5. For SanWen \cite{CRE}, fine-tuning was performed for 10 epochs with a learning rate of 5e-4. The batch size was set to 4, and the gradient accumulation steps were 5e-5. All experiments were conducted on a single NVIDIA A40 GPU (48 GB), and the results are averages from five experiments with different random seeds.

\begin{table}[h!t]
\centering
\begin{tabular}{lcc|cccccccc}
\toprule
\multirow{2}{*}{\textbf{Method}} & \multicolumn{2}{c|}{\textbf{FinRE}} & \multicolumn{2}{c}{\textbf{SanWen}} \\
                        & Dev         & Test        & Dev          & Test        \\
\midrule
T5-base                 & -           & 54.93       & -            & -           \\
ERNIE   2.0             & 63.33       & 60.60       & 79.92        & 77.97       \\
FinBERT-base            & -           & 55.33       & -            & -           \\
ERNIE   3.0             & 64.87       & 62.88       & 81.32        & 82.59       \\
Mengzi-BERT-base-fin    & -           & 58.25       & -            & -           \\
ERNIE   3.0 TITAN       & 65.27       & 63.15       & \textbf{83.07}        & 82.70       \\
BBT-FinT5-base          & -           & 60.62       & -            & -           \\
BBT-FinT5-large         & -           & 61.88       & -            & -           \\
Bert-PAGG               & -           & 53.01       & -            & 73.83       \\
BERT+MultiView          & -           & 53.89       & -            & 72.98       \\ \hline \midrule
\textbf{CRE-LLM(ours)}           & \textbf{69.17}       & \textbf{67.37}       & 81.30         & \textbf{82.74} \\ 
\bottomrule
\end{tabular}
\caption{The F1 score comparison of CRE-LLM with other baselines on FinRE \protect\cite{CRE} and SanWen \protect\cite{CRE} datasets. The results are mainly taken from their original paper. For our proposed CRE-LLM, we display the results of the best setup on FinRE and SanWen. The results of the FinRE and SanWen datasets in RQ1 are all attained utilising Baichuan2-13B, which is fine-tuned with LoRA. The best results in each metric are in \textbf{bold}.}

\label{tab:CRE-LLM}
\end{table}
\subsection{Main Result (RQ1)}
For the DSCRE task, to validate that CRE-LLM outperforms other CRE methods as proposed in this paper, we conducted inference tests through designed experiments. The experimental results, listed in Table \ref{tab:CRE-LLM}, include fine-tuning Baichuan2-13B \cite{Baichuan2} on the FinRE \cite{CRE} and SanWen \cite{CRE} using LoRA \cite{LoRA}. Additionally, we compared CRE-LLM with other baseline models. From Table \ref{tab:CRE-LLM}, it can be observed that CRE-LLM outperforms the majority of baseline models on the SanWen \cite{CRE} and demonstrates significant improvements over all existing CRE methods on the FinRE \cite{CRE}. Comparing with the previous best scores, CRE-LLM increased F1 score on the FinRE \cite{CRE} validation set and test set by approximately 5.98\% and 6.68\%, respectively. This reflects the state-of-the-art performance of CRE-LLM in domain-specific CRE capabilities.

\subsection{Effectiveness of LLM’s Fine-Tuning (RQ2)}
To validate the effectiveness of fine-tuning LLMs, we chose to use the FinRE dataset \cite{CRE}. We randomly selected 20\%, 40\%, 60\%, and 80\% of the training data and fine-tuned Baichuan2-13B \cite{Baichuan2} using LoRA \cite{LoRA} on each subset. We then compared their inference test results with the results of the model fine-tuned on the complete dataset. 
\begin{table}[h!t]
\centering
\setlength{\tabcolsep}{2.1mm}
\begin{tabular}{lccc}
\toprule
\textbf{Fine-Tuning}   \textbf{Setting}                                                               & \textbf{Precision}      & \textbf{Recall}         & \textbf{F1 score}       \\ \midrule
\begin{tabular}[c]{@{}l@{}}Baichuan2-13B\\\ \ \ \ \textit{+20\%Training Data}\end{tabular}          & 63.25          & 60.04          & 61.10           \\
\begin{tabular}[c]{@{}l@{}}Baichuan2-13B\\\ \ \ \ \textit{+40\%Training Data}\end{tabular}          & 66.97          & 63.74          & 64.73          \\
\begin{tabular}[c]{@{}l@{}}Baichuan2-13B\\\ \ \ \ \textit{+60\%Training Data}\end{tabular}          & 69.44          & 66.25          & 67.20           \\
\textbf{\begin{tabular}[c]{@{}l@{}}Baichuan2-13B\\\ \ \ \ \textit{+80\%Training Data}\end{tabular}} & \textbf{69.71} & \textbf{66.89} & \textbf{67.65} \\ \midrule
Baichuan2-13B                                                                       & 69.43          & 66.65          & 67.37          \\ \bottomrule
\end{tabular}
\caption{Ablation study for LLM’s Fine-Tuning on FinRE.}
\label{Fine-Tuning}
\end{table}

As shown in the Table \ref{Fine-Tuning}, the performance of the model on domain-specific CRE tasks gradually improves with an increase in the training data, demonstrating the effectiveness of fine-tuning. Additionally, we observed that when using only 40\% of the training data for fine-tuning, the F1 score already surpassed the original best performance. This indicates that fine-tuning allows LLMs to learn effectively from a limited dataset, achieving commendable performance. Moreover, when fine-tuning with 60\% or more of the training data, there is minimal improvement in the F1 score. This suggests that instruction fine-tuning exhibits good generalization capabilities, emphasizing the importance of the quality of training samples over quantity.

\subsection{The Impact of Instruction Design (RQ3)}
To validate the effectiveness and rationality of the dataset constructed for the DSCRE task using our proposed fine-tuning-based generative approach with open-source LLMs, we conducted ablation experiments on various components. Table \ref{tab:Instruction} shows the effectiveness of the different components of Instruction Design on FinRE \cite{CRE}.
\begin{table}[h!t]
\centering
\setlength{\tabcolsep}{1.3mm}
\begin{tabular}{lccc}
\toprule
\textbf{Fine-Tuning Setting}    & \textbf{Precision}      & \textbf{Recall}         & \textbf{F1 score}       \\ \midrule
Baichuan2-13B w/o EM    & 61.09          & 61.54          & 61.24          \\
Baichuan2-13B w/o AT    & 68.67          & 65.13          & 66.23          \\
\textbf{Baichuan2-13B w/o TR}    & 69.28          & \textbf{67.07} & \textbf{67.50} \\
Baichuan2-13B w/o AT+TR & 67.44          & 64.42          & 65.22          \\ \midrule
Baichuan2-13B           & \textbf{69.43} & 66.65          & 67.37          \\ \bottomrule
\end{tabular}
\caption{Ablation study for Instruction Design on FinRE.}
\label{tab:Instruction}
\end{table}

\textbf{Effectiveness of Using ``[]" Entity Markers (EM).} As shown in Tabel \ref{tab:Instruction}, the model when fine-tuned without utilizing ``[]" entity markers in the dataset resulted in a decrease in Precision, Recall and F1 score by 13.65\%, 8.30\% and 10.11\% respectively. Using ``[]" entity markers proves to be effective. It can facilitate LLMs in locating and understanding the given entities in the sentence, as well as aligning entities during the inference process. This helps improve the performance of CRE .

\textbf{Effectiveness of Adding Triplets (AT) After the Input Text} Tabel \ref{tab:Instruction} demonstrates that removing the triplet part leads to a decrease in Precision, Recall, F1 score by 1.11\%, 2.33\% and 1.81\% respectively. The absence of the triplet in the input text during the fine-tuning phase makes the model only observe the positional relation between entities from the output part. It hinders the learning process and reducing the performance of CRE. 

\textbf{Effectiveness of the Triplet Results(TR)} As shown in Table \ref{tab:Instruction}, from the results of the ablation experiments, it can be observed that TR does not significantly improve Precision, Recall and F1 Score. However, in the absence of AT, TR can respectively increase Precision, Recall and F1 Score by 2.95\%, 3.46\%, and 3.39\%. This is because in such a scenario, demanding direct output of relation extraction results from LLM may lead to insufficient understanding of entities and their relation reasoning. The setting of TR provides partial informative cues and aligns with entities indicated in the input, leveraging the well-established learning patterns of LLM. This allows for better understanding of task requirements, resulting in superior performance on domain-specific CRE tasks. Therefore, we consider TR to be effective, contributing to enhanced robustness of LLM in domain-specific CRE performance.

\textbf{Comparison among all components.} The results of ablation experiments for all components are shown in Table \ref{tab:Instruction}, where the settings of EM and AT significantly improve Precision, Recall and F1 score. Particularly, EM has the most significant impact on the performance of the model in CRE. For the TR part, its influence on the performance of the model is substantial without AT settings. However, with AT settings, its impact becomes very limited. Consequently, AT has a stronger effect on LLM's performance in CRE than TR, with TR playing a role similar to AT but to a lesser extent. In summary, EM has the most significant impact on LLM's performance in DSCRE, followed by AT, and TR has the smallest impact.

\subsection{Comparison with ChatGPT (RQ4)}
To illustrate why CRE-LLM chooses to fine-tune open-source LLMs for relation extraction, instead of directly using ChatGPT to accomplish this task, experiments were designed to compare these two approaches. Therefore, we conducted experiments on FinRE \cite{CRE} for relation extraction. Table \ref{tab:CRE-LLM and GPT} shows the Comparison with ChatGPT\cite{ChatGPT}.
\begin{table}[h!t]
\centering
\setlength{\tabcolsep}{1.2mm}
\begin{tabular}{lccc}
\toprule
\textbf{Method}                                                                     & \textbf{Precision}      & \textbf{Recall}         & \textbf{F1 score}       \\ \midrule
\begin{tabular}[c]{@{}l@{}}ChatGPT\\\ \ \ \ \textit{+Classify-then-Extract}\end{tabular}   & 27.59          & 25.86          & 26.44          \\ 
\begin{tabular}[c]{@{}l@{}}ChatGPT\\\ \ \ \ \textit{+Generate-then-Retrieval}\end{tabular} & 25.86          & 24.14          & 24.71          \\  \hline \midrule
\textbf{Baichuan2-13B+LoRA}                                                         & \textbf{69.43} & \textbf{66.65} & \textbf{67.37} \\ \bottomrule
\end{tabular}
\caption{CRE result comparison of CRE-LLM with ChatGPT on FinRE. For our proposed CRE-LLM, we display the results of the basic setup on FinRE. The best results in each metric are in \textbf{bold}.}
\label{tab:CRE-LLM and GPT}
\end{table}

\textbf{Comparison with Classify-then-Extract.} This method involves constructing a Relation Set containing all distinguished relations from the RE dataset. The prompt is modified to introduce the Relation Set, which allows ChatGPT \cite{ChatGPT} to select the appropriate relation between the given entities from the Relation Set. As shown in Table \ref{tab:CRE-LLM and GPT}, despite ChatGPT \cite{ChatGPT} having a larger number of model parameters, it is not open-source and cannot be fine-tuned, posing a challenge for generating standard relation extraction results directly. The challenge presented by the dataset is the necessity to provide 44 distinguished relations for ChatGPT \cite{ChatGPT} to differentiate. This necessitates the input prompt being longer and more challenging for the model to understand, consequently reducing the performance of CRE.

\textbf{Comparison with Generate-then-Retrieval.} This method involves using ChatGPT \cite{ChatGPT} to directly infer and generate relations. In contrast to the Classify-then-Extract approach, this method does not necessitate the input of all special relations, simplifying the prompt significantly. However, the generated results are more diverse. SimCSE \cite{SimCSE} is employed to align the results with the Relation Set for relation extraction. As shown in Table \ref{tab:CRE-LLM and GPT}, due to the lack of fine-tuning, ChatGPT \cite{ChatGPT} exhibits weaker internal understanding and logical reasoning capabilities for domain-specific Chinese text. It fails to generate results with precise meanings. Consequently, the performance of relation extraction is naturally not ideal after retrieval alignment through SimCSE \cite{SimCSE}.

\subsection{Analysis of Efficiency (RQ5)}
For domain-specific CRE tasks, training efficiency and costs also require significant attention. To illustrate that CRE-LLM achieves higher efficiency and lower cost, our experimental results as shown in Table \ref{tab:CRE-LLM and ERNIE 3.0}.
\begin{table}[h!t]
\setlength{\tabcolsep}{0.02mm}
\begin{tabular}{lcc}
\toprule
\textbf{Method}                                                                          & \textbf{Trainable params↓}                                    & \textbf{Training Time↓}                                       \\ \midrule
\begin{tabular}[c]{@{}l@{}}ERNIE 3.0\\\ \ \ \ \textit{+Progressive Learing}\end{tabular}        & \multirow{1}{*}{100M}  & \begin{tabular}[c]{@{}l@{}}11h30m\\ 4h\end{tabular} \\ \hline \midrule
\begin{tabular}[c]{@{}l@{}}\textbf{Baichuan2-13B+LoRA}\\\ \ \ \ \textit{\textbf{+40\%Training Data}}\end{tabular} & \multirow{2}{*}{\textbf{0.0655M}}                            & \textbf{1h08m}                                               \\
Baichuan2-13B+LoRA                                                              &                                                     & 2h32m                                               \\ \bottomrule
\end{tabular}
\caption{Comparison of CRE-LLM with ERNIE 3.0 on FinRE. }
\label{tab:CRE-LLM and ERNIE 3.0}
\end{table}

The specific experimental results are presented in Table \ref{tab:CRE-LLM and ERNIE 3.0}. Although ERNIE 3.0 \cite{ERNIE3.0} exhibits superior performance after fine-tuning on the SanWen \cite{CRE}, Table \ref{tab:CRE-LLM and ERNIE 3.0} reveals that the number of parameters and GPU memory consumption required for general NLP tasks after fine-tuning ERNIE 3.0 is significantly more than for CRE-LLM. For instance, during the fine-tuning phase, ERNIE 3.0 would need at least eight 32GB V100 GPUs, indicating higher requirements for environmental configuration. Additionally, the fine-tuning training duration is also substantially longer. In contrast, CRE-LLM, which employs PEFT \cite{peft} framework, achieves efficient parameter fine-tuning for LLMs. This results in a significant reduction in GPU memory usage and a lowering of environmental configuration demands, thereby enabling more general teams and projects to adopt it.

\subsection{Error Analysis (RQ6)}

We analyzed instances in the FinRE \cite{CRE} test set where CRE-LLM failed to achieve correct CRE, and conducted a statistical analysis of the errors. The specific findings are summarized as follows:

\textbf{Entity Relation Understanding Errors (52.21\%).} The primary source of errors in relation extraction is a misunderstanding of relations between entities. This is mainly due to the inherent difficulty of relation extraction in the dataset and the model's limited ability to precisely infer relations between entities in domain-specific natural language texts.

\textbf{Errors with Multiple Relations between Entities (26.69\%).} Another category of errors arises when there are multiple relations between entities. For instance, in sentences like ``With the establishment of [Ant Financial], [Alibaba]'s layout in the financial business has been officially clarified," , and the relations of the given entities are ``ownership" and ``establishment". The model may overlook one of the relations while generating the relation extraction result, leading to lower the performance of CRE.

\textbf{``NA" Relation Errors (20.47\%).} The FinRE \cite{CRE} dataset contains a special ``NA" relation that is challenging to express in the text. Even for general readers, understanding the relation between the given entities may be difficult. This complexity poses a challenge for the model to achieve accurate CRE.

\textbf{Nonexistent Relation Errors (0.622\%).} Since CRE-LLM does not provide any relation options and relies on learning from training data to obtain the Relation Set of the dataset. Consequently, there may be instances where the model, despite fine-tuning, generates relation extraction results that do not exist in the Relation Set. This is due to the fine-tuned LLMs still retain certain independent capabilities in text understanding, generation, and generalization.

\section{Conclusion}

In this work, we propose CRE-LLM, a large language model framework for domain-specific Chinese relation extraction (DSCRE) based on fine-tunned open-source large models. This method represents a significant shift from traditional approaches. It employs the PEFT \cite{peft} framework and open-source LLMs with numerous parameters to achieve a simple and efficient end-to-end generative relation extraction. It addresses inherent challenges such as complex network structure design, poor perception and high consumption of fine-tuning. 

Our experimental results, based on two standard domain-specific CRE benchmarks, namely FinRE \cite{CRE} and SanWen \cite{CRE}, demonstrate that CRE-LLM achieves state-of-the-art(SOTA) performance on the DSCRE tasks. Moreover, the simplicity, flexibility, and especially the efficiency of our framework make it a promising direction for applying LLMs to DSCRE tasks that involve stronger domain specificity and more challenging semantic understanding.

\section*{Acknowledgment}
This work is supported in part by BUPT Excellent Ph.D. Students Foundation (No. CX2023133).
\bibliographystyle{named}
\bibliography{ijcai24}

\end{document}